\documentclass{article}

\usepackage[final]{neurips_2022}

\usepackage[utf8]{inputenc} 
\usepackage[T1]{fontenc}    
\usepackage{hyperref}       
\usepackage{url}            
\usepackage{booktabs}       
\usepackage{amsfonts}       
\usepackage{nicefrac}       
\usepackage{microtype}      
\usepackage{xcolor}         

\usepackage{amsmath}
\usepackage{amssymb}
\usepackage{mathtools}
\usepackage{amsthm}

\usepackage[capitalize,noabbrev]{cleveref}

\usepackage[normalem]{ulem}

\usepackage{multicol}
\usepackage{multirow}
\usepackage{wrapfig}

\usepackage{subcaption}
\usepackage{graphicx}

\usepackage[textsize=tiny]{todonotes}
\setcitestyle{authoryear,round}

\title{Recurrent Memory Transformer}

\author{%
  Aydar Bulatov$^1$\\
  \texttt{bulatov.as@phystech.edu} \\
  \And
  Yuri Kuratov$^{1,2}$\\
  \texttt{yurii.kuratov@phystech.edu} \\
  \And
   Mikhail S.~Burtsev$^{1,2}$\\
  \texttt{burtcev.ms@mipt.ru} \\
  \AND
  \\
  $^1$Neural Networks and Deep Learning Lab,\\
  Moscow Institute of Physics and Technology, Dolgoprudny, Russia \\
  $^2$AIRI, Moscow, Russia
}

\begin{document}

\maketitle

\begin{abstract}
  Transformer-based models show their effectiveness across multiple domains and tasks. The self-attention allows to combine information from all sequence elements into context-aware representations. However, global and local information has to be stored mostly in the same element-wise representations. Moreover, the length of an input sequence is limited by quadratic computational complexity of self-attention.
  In this work, we propose and study a memory-augmented segment-level recurrent Transformer (RMT). Memory allows to store and process local and global information as well as to pass information between segments of the long sequence with the help of recurrence.
  We implement a memory mechanism with no changes to Transformer model by adding special memory tokens to the input or output sequence. Then the model is trained to control both memory operations and sequence representations processing.
  Results of experiments show that RMT performs on par with the Transformer-XL on language modeling for smaller memory sizes and outperforms it for tasks that require longer sequence processing. We show that adding memory tokens to Tr-XL is able to improve its performance. This makes Recurrent Memory Transformer a promising architecture for applications that require learning of long-term dependencies and general purpose in memory processing, such as algorithmic tasks and reasoning.
\end{abstract}

\section{Introduction}
\label{introduction}

\begin{wrapfigure}{r}{0.45\linewidth}
\includegraphics[width=\linewidth]{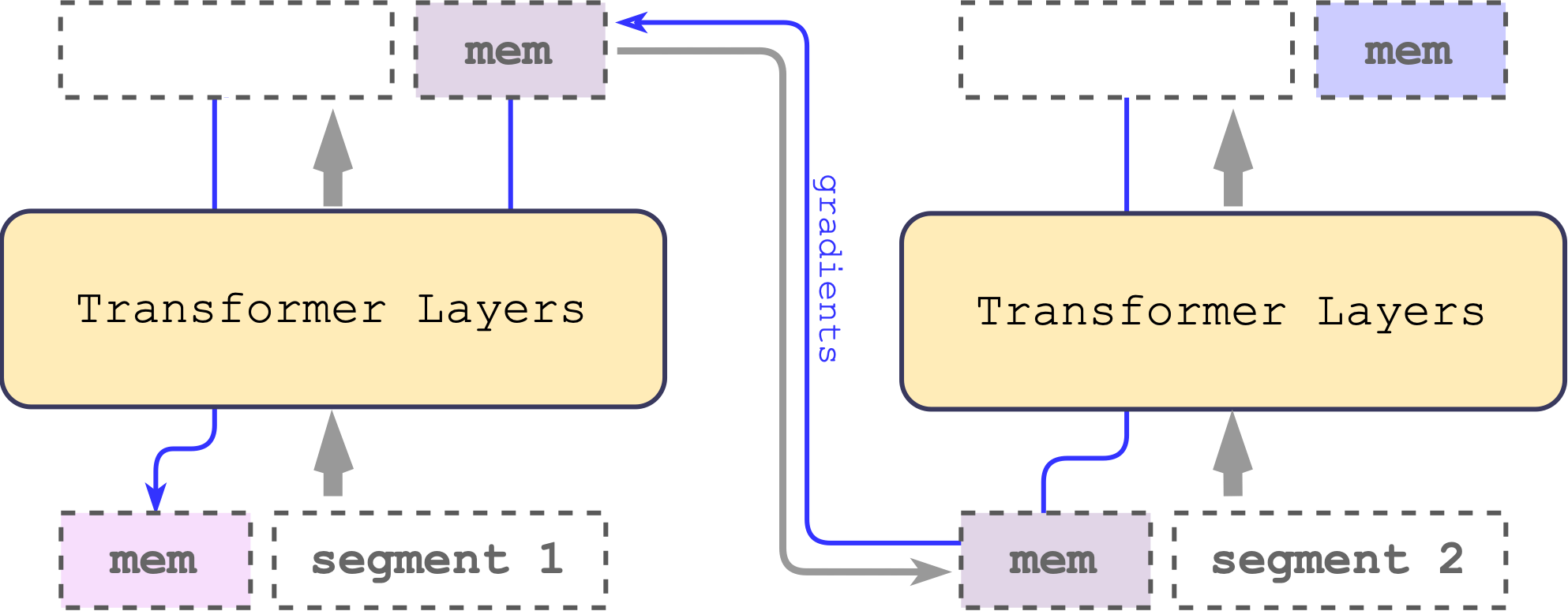}
\caption{\small\textbf{Recurrent Memory Transformer.} Memory is added as tokens to the input sequence and memory output is passed to the next segment. During training gradients flow from the current segment through memory to the previous segment.}
\label{RMT_simple}
\end{wrapfigure}

Transformers~\citep{vaswani2017attention} have been widely adopted across multiple domains and tasks~\citep{RadfordNarasimhanEtAl_2018_Improving_language_understanding_by_generative_pre-training, speech-transformer, devlin2019bert, dosovitskiy2021vit, dalle, jaegle2021perceiverio}.
The key component of Transformer layer is a self-attention. Self-attention allows to update each sequence element representation  with information from all other elements in the sequence. As a result, rich contextual representation for every element is generated at the end of encoding. This way, global sequence-level and local information are stored in a single representation. However, this mixing of two types of information in a single representation has limitations. Distributed storage of global features across all sequence elements results in global features "blurring" and makes it harder to access them. Another well-known deficiency of Transformers is poor scaling of self-attention with input sequence length that hurts its applications to long inputs~\citep{child2019generating, guo2019startransformer, dai2019transformerxl, beltagy2020longformer, ainslie2020encoding, zaheer2020big, wang2020linformer, choromanski2020performer}.

Our work introduces a memory-augmented segment-level recurrent Transformer named Recurrent Memory Transformer (RMT). RMT uses a memory mechanism based on special memory tokens~\citep{burtsev2020memory-transformer} added to the input sequence. Memory tokens provide additional reserved capacity to the model that could be used to process information which is not directly representing any element in the input sequence. To process long sequences, we split them into segments and pass memory states from a previous to a current segment. This memory passing makes the model recurrent and removes the input sequence length limitations. RMT model can theoretically work with infinite lengths but, in practice, it is limited by memory capacity and the efficiency of memory access/update operations. Our implementation of both memory and recurrence in RMT requires no changes to the Transformer model because modifications are made only to the input and output sequences of the model.

We tested RMT on the tasks that require global information about the whole input sequence to be solved. We use copy, reverse, and associative retrieval tasks in the setting where the input sequence is split into segments. RMT and Transformer-XL perfectly solve these tasks, but exceeding some value of sequence length, RMT starts to outperform Transformer-XL. Also, we experimentally show that the proposed Recurrent Memory Transformer requires less memory size to perform closely to Transformer-XL on language modeling tasks. RMT code and experiments are available\footnote{\url{https://github.com/booydar/LM-RMT}. The code, results of the raw experiments and hyperparameters are provided in the supplementary materials and on GitHub.}.

\textbf{Contributions}

1. In this study we augment Transformer with token based memory storage and segment-level recurrence.

2. We experimentally evaluate proposed architecture as well as vanilla Transformer and Transformer-XL on memory-intensive tasks such as copy, reverse, associative retrieval, and language modeling. We show that RMT outperforms Transformer-XL for sequence processing tasks and on par with Transformer-XL on language modeling but requires less memory.

3. We show that Tr-XL cache could be combined with RMT leading to better performance on language modeling.

4.  We analysed how the Transformer model learns to use  memory. Specific interpretable memory read-write patterns of attention are shown.

\section{Related work}

In our study we add a memory to general purpose attention based neural architecture. Memory is a recurrent topic in neural networks research. It had started from the early works ~\citep{mcculloch1943logical,stephen1956kleene} and significantly progressed in 90's with introduction of \textit{Backpropagation Through Time} learning algorithm \citep{werbos1990backpropagation} and \textit{Long-Short Term Memory} (LSTM) \citep{10.1162/neco.1997.9.8.1735} neural architecture. Today memory-augmented neural networks (MANNs) usually rely on some kind of recurrent external-memory which is separate  from the model's parameters. \textit{Neural Turing Machines }(NTMs) \citep{graves2014neural} and \textit{Memory Networks }\citep{weston2014memory} are equipped with a storage for vector representations that can be accessed with an attention mechanism. Memory Networks \citep{weston2014memory,sukhbaatar2015endtoend} were designed to enable reasoning by sequential attention over to the content of a memory. 
NTMs followed by \textit{Differentiable Neural Computer} (DNC) \citep{graves2016hybrid} and \textit{Sparse DNC} \citep{rae2016scaling} are implemented as recurrent neural networks able to write to memory storage over time. All these models are differentiable and can be trained via backpropagation through time (BPTT). Parallel line of research extends recurrent neural networks such as LSTM with data structures like stacks, lists, or queues~\citep{arm2015inferring,grefenstette2015learning}. MANN architectures with a more advanced addressing mechanisms such as address-content separation and multi-step addressing were proposed in \citep{gulcehre2016dynamic,gulcehre2017memory,meng2018context}. The Global Context Layer model \citep{meng2018context} uses the idea of address-content separation to solve the difficulty of training content-based addressing in the canonical NTM.

The recent rise of Transformer models also resulted in introduction of a number of new memory architectures.
\textit{Transformer-XL}~\citep{dai2019transformerxl} introduces a segment-level recurrence at the level of hidden representations. These representations of a sequence are computed and stored in the cache to be reused as an extended context for the next segment. \textit{Compressive Transformer} \citep{rae2019compressive} adds the second layer of memory to Transformer-XL. This memory compresses and stores information from the cache. \textit{$\infty$-former}~\citep{martins2021infinity} utilizes continuous-space attention and represents input sequence as a continuous signal to make long-term memory unbounded. \textit{Memory Layers} \citep{lample2019large} model has a product key memory layer instead of a feed-forward layer within Transformer block to increase model capacity.

In many variations of Transformer different sorts of global representations are added. Among them are \textit{Star-Transformer} \citep{guo2019startransformer}, \textit{Longformer} \citep{beltagy2020longformer}, \textit{GMAT}~\citep{gupta2020gmat}, \textit{Extended Transformer Construction} (ETC) \citep{ainslie2020encoding} and \textit{Big Bird} \citep{zaheer2020big}. All these architectures re-design self-attention mechanism to reduce it computational complexity with and ensure input coverage with the help of global representations. \textit{Memory Transformer} \citep{burtsev2020memory-transformer} keeps Transformer model intact and adds memory by extending input sequence with special memory tokens.  Perceiver~IO~\citep{jaegle2021perceiverio} maps an entire arbitrary input to the fixed number of latent representations. Transformer layers do further processing over latent memory representations only. 



Segment-level recurrence in Transformers is actively explored in a number of studies. Transformer-XL, Compressive Transformer keep previous states and re-use them in subsequent segments. Ernie-Doc~\citep{ding-etal-2021-ernie-doc} improves processing by using same-layer recurrence instead of attending to previous layer outputs of a precedent segment. Memformer~\citep{wu2020memformer} introduces a dedicated memory module to keep previous hidden states in summarized representations. Memformer uses two special layers added to the Transformer model. Memory cross-attention layer reads from memory and memory slot attention layer updates it. MART~\citep{lei2020mart} has a similar approach as Memformer but uses memory update rules analogous to LSTM~\citep{10.1162/neco.1997.9.8.1735} and GRU~\citep{cho2014gru}. FeedBack Transformer~\citep{fan2020feedback-transformer} goes further with full, and not segment-level, recurrence. FeedBack Memory merges past hidden representations from all layers into a single vector and makes it accessible to the computations at any layer. The disadvantage of full recurrence is that it is less parallelizable. FeedBack Memory requires every sequence element to be processed sequentially. In segment-level recurrent models, all elements of a segment are processed by Transformer layers in parallel. Only segments are processed sequentially. Staircase Transformer~\citep{ju2021staircase} combines segment-level recurrence and depth recurrence. Staircase models use the output for previous segments and pass them as input for the next segment. Our Recurrent Memory Transformer is based on special memory tokens similar to Memory Transformer, segment-level recurrence as in Transformer-XL, and depth-recurrent mechanism for memory processing similar to Staircase.

\section{Recurrent Memory Transformer}


Transformer-XL~\citep{dai2019transformerxl} extends Transformer model with state re-use cache mechanism for segment-level recurrence and relative position encoding. Input sequence is split on segments processed sequentially. Hidden states computed for the previous segment $M^{n}$ are cached for each transformer layer $n$. The input of the layer $n$ consists of the last $m$ states from the cached memory and output of previous Transformer layer for the current segment $\tau$:
\begin{equation*}
\tilde{H}^{n-1}_{\tau} = [ SG(M^{n-1}_{-m:}) \circ H^{n-1}_\tau ],
\end{equation*}
here, SG stands for stop-gradient, $\circ$ denotes concatenation. Cached states allow to increase effective context size of Transformer model and save on compute operations.

\begin{figure*}[h]
\vskip 0.0in
\begin{center}
\centerline{\includegraphics[width=1.0\linewidth]{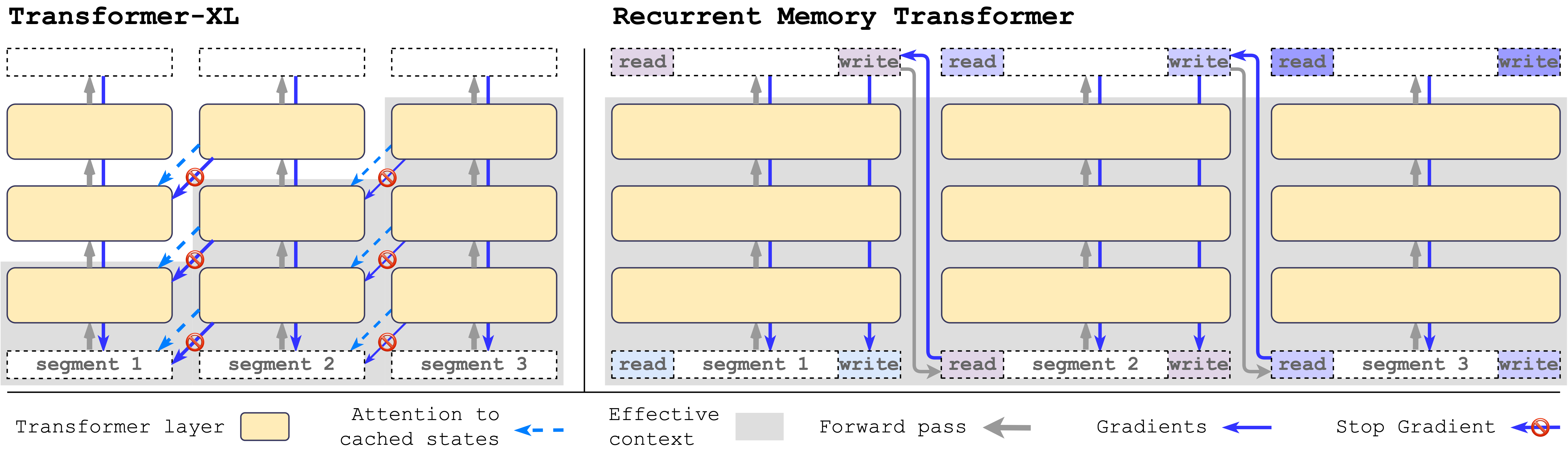}}
\caption{\small\textbf{Comparison of Recurrent Memory Transformer (RMT) and Transformer-XL architectures.} Recurrent Memory Transformer augments Transformer with global memory tokens and passes them to allow a segment-level recurrence. Special read/write memory tokens are added to the input sequence. Multiple memory tokens can be used in each read/write block. Updated representations of write memory are passed to the next segment. During training, RMT uses BPTT to propagate gradient to previous segments through memory tokens representation. Effective context length for recurrence with memory is not limited by the depth of a network which is the case for the cache of Transformer-XL.}
\label{RMT_full}
\end{center}
\vskip -0.4in
\end{figure*}

Then, $\tilde{H}^{n-1}_{\tau}$ goes to Transformer layer $TL$ to produce layer $n$ outputs for segment $\tau$:
\begin{equation*}
    \begin{gathered}
    H^{n}_{\tau} = TL(Q^n_{\tau}, K^n_{\tau}, V^n_{\tau}),
    Q^n_{\tau} = W^n_q H^{n-1}_{\tau}; K^n_{\tau} = W^n_k \tilde{H}^{n-1}_{\tau}, V^n_{\tau} = W^n_v \tilde{H}^{n-1}_{\tau}.
    \end{gathered}
\end{equation*}
In Transformer-XL, self-attention layers are modified to use relative position encodings to improve generalization to longer attention lengths. The overall architecture is shown in the~\cref{RMT_full}.

Memory augmented Transformers such as GMAT, ETC, Memory Transformer~\citep{gupta2020gmat, ainslie2020encoding, burtsev2020memory-transformer} proposed to use special global tokens as storage for representations. Usually, memory tokens are added to the beginning of the input sequence. However, in decoder-only architectures  the causal attention mask makes impossible for memory tokens at the start of the sequence to collect information from the subsequent tokens. On the other hand, if memory tokens are placed at the end of the sequence then preceding tokens unable to access their representations. To solve this problem we add a recurrence to the sequence processing. Representations of memory tokens placed at the end of the segment are used as an input memory representations at the start as well as at the end of the next segment.

In the Recurrent Memory Transformer  input is augmented with special \texttt{[mem]} tokens, processed in a standard way along with the sequence of tokens. Each memory token is a real-valued vector. $m$ memory tokens are added at the beginning of the segment tokens representations $H_{\tau}^0$ and the same $m$ tokens are added at the end:
\begin{equation*}
\tilde{H}^{0}_{\tau} = [H_{\tau}^{mem} \circ H_{\tau}^0 \circ H_{\tau}^{mem}],
\bar{H}^N_\tau = \text{Transformer}(\tilde{H}^{0}_{\tau}),
[H_{\tau}^{read} \circ H_{\tau}^N \circ H_{\tau}^{write}] := \bar{H}^N_\tau,
\end{equation*}
here $N$ is a number of Transformer layers.

The starting group of memory tokens functions as a read memory that allows sequence tokens to attend to memory states produced at the previous segment. The ending group works as a write memory that can attend to all current segment tokens and update representation stored in the memory. As a result, $H_{\tau}^{write}$ contains updated memory tokens for the segment $\tau$.

Segments of the input sequence are processed sequentially. To enable recurrent connection between segments, we pass outputs of the memory tokens from the current segment to the input of the next segment:
\begin{equation*}
\begin{gathered}
H_{\tau+1}^{mem} := H_{\tau}^{write},
\tilde{H}^{0}_{\tau+1} = [H_{\tau+1}^{mem} \circ H_{\tau+1}^0 \circ H_{\tau+1}^{mem}].
\end{gathered}
\end{equation*}

Both memory and recurrence in the RMT are based only on global memory tokens. It allows to keep the backbone Transformer unchanged and make RMT memory augmentation compatible with any model from the Transformer family.
Memory tokens operate only on the input and output of the model. In this study we implement RMT on top of the original Transformer-XL code. Both architectures are shown in~\cref{RMT_full}.

Recurrence in the RMT is different compared to the Transformer-XL because the former stores only $m$ memory vectors per segment. On the other hand, the Transformer-XL stores $m \times N$ vectors per segment. Also, in the RMT model memory representations from the previous segment are processed by Transformer layers together with the current segment tokens. This makes memory part of RMT effectively deeper in a number of applied Transformer layers $\tau \times N$. Additionally, we allow all memory tokens in the read/write block to access all other tokens in the same block. The causal attention mask is applied only to tokens of the input sequence (\cref{fig:mem_operations}(d)).

We train the RMT with Backpropagation Through Time (BPTT). During backward pass, unlike in Transformer-XL, memory gradients are not stopped between segments. The number of previous segments to backpropagate is a hyperparameter of a training procedure. We vary BPTT unroll in our experiments from 0 to 4 previous segments. Increasing this parameter is computationally expensive and requires a lot of GPU RAM. However, techniques such as gradient checkpointing could be used to alleviate this problem.


\section{Experiments}
\label{sec:experiments}

We designed our experiments to evaluate the ability of Recurrent Memory Transformers to preserve long-term dependencies across multiple input segments. The first set of experiments includes copy, reverse, associative retrieval, and quadratic equations tasks. The second one addresses language modeling task for word-level on WikiText-103~\citep{merity2017wikitext} and for character-level on enwik8~\citep{enwik8}. We compare Recurrent Memory Transformer with Transformer and Transformer-XL models.

Our RMT implementation is based on Transformer-XL repository\footnote{\url{https://github.com/kimiyoung/transformer-xl}}. 
The full set of hyperparameters is available in our repository as well as in supplementary materials. Language modeling experiments follow the same model and training hyperparameters as Transformer-XL. WikiText-103 experiments use 16-layer Transformers (10 heads, 410 hidden size, 2100 intermediate FF), enwik8 -- 12 layer Transformers (8 heads, 512 hidden size, 2048 intermediate FF). We used Adam optimizer~\cite{adam} with linear schedule learning rate starting from $0.00025$ for 200,000 steps for WikiText-103 and 400,000 steps for enwik8. We refer to Transformer-XL with memory size equal to zero as a Baseline.
With this experimental setup we were able to reproduce  results for the Transformer-XL model close to the original paper.

\textbf{Algorithmic Tasks.} We evaluate RMT on algorithmic tasks that require information about the whole input sequence to be solved successfully. In a recurrent setting, the model has to keep information about all previous segments to make predictions.

In the \textit{Copy} task, an input sequence should be replicated twice after a special start-to-generate token. In the \textit{Reverse} task, an input sequence should be generated in a reverse order. Input for the \textit{Associative Retrieval} task consists of $N$ key-value pairs. Then one key is randomly selected, and the task is to produce an appropriate value for the selected key. Another task is to solve quadratic equations. One example consists of an equation, its solution with discriminant, and an answer. The task is to generate a solution and answer, while only answer quality is evaluated.

For all tasks, input and output sequences are split into segments and processed by models sequentially. Datasets for algorithmic tasks were randomly pre-generated, the same data was used in all experiments, and character-level tokenization was used. Because Transformer-XL and RMT are decoder-only Transformer models, we don't  compute loss over the input sequence before the start-to-generate token. The loss is computed over target sequence segments only (see \cref{app:alg_training_details} for details).



\textbf{Language Modeling and NLP.} We use two standard benchmarks for language modeling: WikiText-103 and enwik8. WikiText-103~\citep{merity2017wikitext} is used for word-level language modeling and contains 103M words from English Wikipedia articles. Enwik8~\citep{enwik8} is used for character-level and consists of $10^8$ first bytes of XML text dump of the English Wikipedia. Vocabulary contains 267735 words and 204 characters for Wikitext-103 and enwik8 tokenizers accordingly.

We compare Recurrent Memory Transformer with decoder-only Transformer and Transformer-XL as baselines. Model size and training parameters are selected to match Transformer-XL paper. For Wikitext-103 an input context length was set to 150 tokens, and for enwik8 it was set to 512 characters.
Another set of experiments inspected how RMT handles long-term dependencies and recurrence. We increased the number of segments and recurrent steps by making segments smaller (50 tokens for WikiText-103, 128 characters for enwik8). The increased number of recurrent steps makes language modeling tasks harder for RMT because information has to be stored in the same amount of memory for more steps.

As a testbed for the real-life application scenario we select popular long-text classification benchmark Hyperpartisan news \citep{kiesel2019semeval}. Instead of pre-training RMT from scratch we add recurrent memory mechanism to the most widely adopted models from HuggingFace Transformers \citep{wolf2020transformers}. Specifically, we augment 500 input tokens of already pretrained BERT-base, RoBERTa-base, DeBERTa-base and T5-base with the recurrent memory of size 10 and fine-tune on the target task.

\section{Results}
\label{results}
\begin{figure*}[h]
\vskip -0.1in
\begin{center}
\centerline{\includegraphics[width=\linewidth]{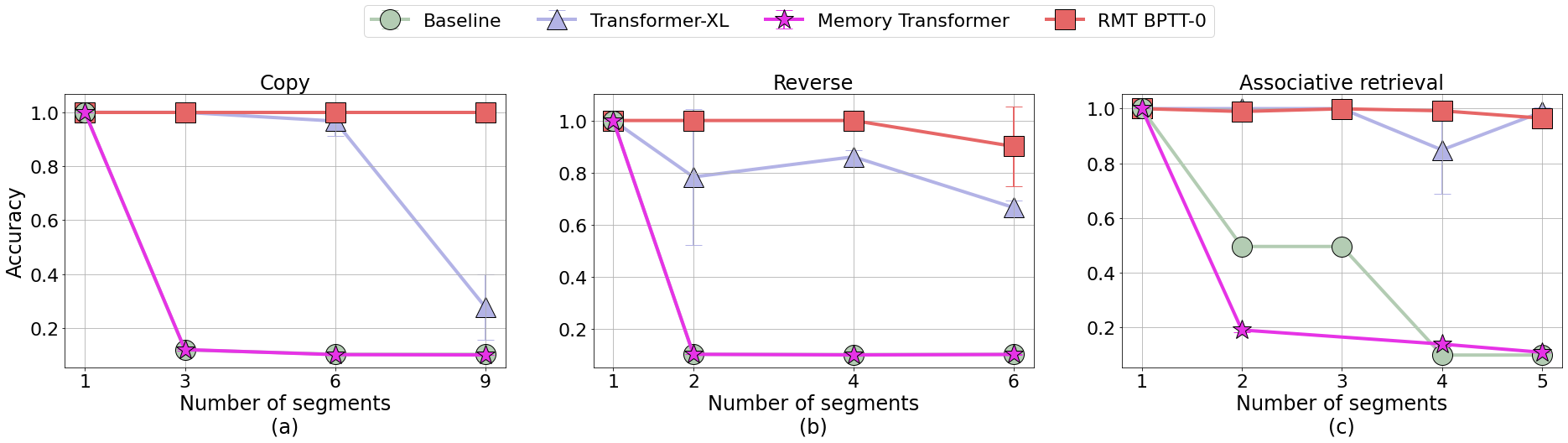}}
\caption{\small\textbf{RMT outperforms Transformer-XL on Copy and Reverse tasks as a number of segments increases.} Panels show test set per-character accuracy on copy, reverse, and associative retrieval tasks (from left to right). Memory/cache size equals to the length of a segment for both models. RMT does not pass gradients between segments in this experiment. MT results are the same as for the Baseline. Source/target sequence lengths for copy, reverse, and associative retrieval tasks: 24/48, 24/24, 10/1.}
\label{fig:algotasks}
\end{center}
\vskip -0.3in
\end{figure*}

Baseline, Transformer-XL (Tr-XL) and RMT perform perfectly in the single segment setting on copy and reverse tasks (\cref{fig:algotasks}). In this case, the models do not need recurrence because the whole sequence is available.
When the number of segments is larger than one, non-recurrent baseline struggles to solve tasks, but both memory models demonstrate ability to retain required information from the previous segments in memory.

On Copy and Reverse tasks as a number of segments increases, RMT starts to outperform Transformer-XL with memory sizes less than the number of all previous tokens. With the number of segments up to 6 mean accuracy of Transformer-XL drops by up to 0.2 points, and with 9 segments plunges close to the baseline without memory. Associative Retrieval results are similar with the number of segments up to 4. RMT manages to solve the task with Transformer-XL closely behind. However, in the setting with 5 segments, RMT performance slightly decreases and Transformer-XL average accuracy rises higher.

\begin{figure}[t]
\begin{center}
\begin{subfigure}[t]{0.45\linewidth}
    \centerline{\includegraphics[width=0.8\linewidth]{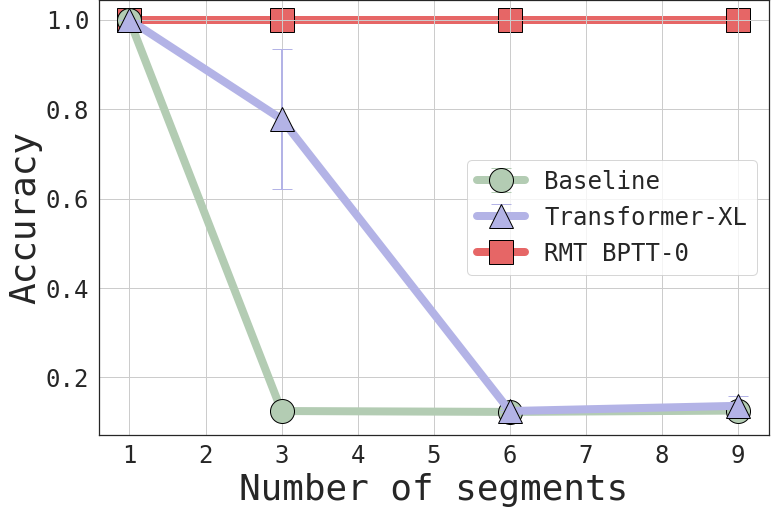}}
    \caption{}
    \label{fig:copy120}
\end{subfigure}
\begin{subfigure}[t]{0.45\linewidth}
    \centerline{\includegraphics[width=0.8\linewidth]{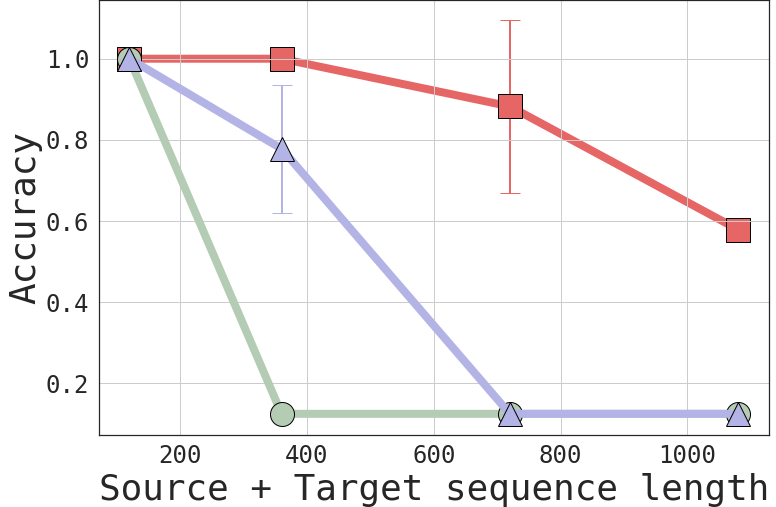}}
    \caption{}
    \label{fig:copy360}
\end{subfigure}\hfill
\caption{\small \textbf{RMT scales better with a number of segments and sequence size.} (a) RMT is able to solve copy task perfectly up to 9 segments for a fixed sequence length of 360, while Tr-XL fails. (b) RMT learns to use memory of the same fixed size (60 tokens) more effectively than TR-XL as a sequence length to copy increases (a segment size is 120 for the both models).}
\label{fig:rmt_seq_analysis}
\end{center}
\vskip -0.45in
\end{figure}

We analyze how a number of segments, sequence length, a length of training context, and memory size affect models' performance on Copy task (\cref{fig:rmt_seq_analysis}). As we split a sequence into more segments it becomes more crucial to be able to pass information between segments. We split 360 tokens of source + target sequence into multiple segments. In~\cref{fig:copy120} we observe that Transformer-XL performance starts to degrade and eventually falls to the baseline model performance as the number of segments increases. In contrast, RMT continues to solve the task perfectly. In a more extreme setting, when we keep memory size fixed, but increase the total length of a sequence to copy Transformer-XL fails shortly, while RMT starts to gradually degrade only after the length of 720 tokens (\cref{fig:copy360}).

On the Quadratic Equations task (\cref{tab:quadratic_eq}) we have checked that it is possible to solve the task with the Transformer baseline and no segmentation used. The baseline in this case defines upper bound for this task. With multiple segments recurrency RMT solves the task perfectly, while Transformer-XL finds the task challenging.

\begin{wraptable}{r}{0.5\linewidth}
\vspace{-0.2in}
\caption{\small \textbf{Quadratic equations task.} Sequence of 180 tokens consists of quadratic equation, a solution, and an answer. It is split into a number of segments with an answer in the last segment. Accuracy equals 1.0 if the full answer is predicted correctly.}
\label{tab:quadratic_eq}
\begin{center}
\begin{small}
\begin{sc}
\fontsize{7}{8}
\selectfont 
\begin{tabular}{lccl}
\toprule
Model          & memory & segments & $\text{Acc}_{\pm \text{std}}$ \\
\midrule
Baseline        &  0           &  1          & 0.99 \tiny{$\pm$ 0.01} \\
Transformer-XL           &  30          &  6          & 0.93 \tiny{$\pm$ 0.02} \\
RMT             &  30          &  6          & 0.99 \tiny{$\pm$ 0.002} \\
\bottomrule
\end{tabular}
\end{sc}
\end{small}
\end{center}
\vskip -0.2in
\end{wraptable}

The results of experiments on word-level language modeling on WikiText-103 are shown in \cref{tab:wt103}. In the first section with a segment length of 150, Tr-XL and RMT outperform the baseline and Memory Transformer (MemTr) by a large margin. It shows the significance of increased effective context length by Tr-XL cache or RMT memory for language modeling.
RMT improves over MemTr memory mechanism with read/write blocks. The best RMT models with memory size 10 and 25 show similar performance as Transformer-XL with a memory size equal to 75. RMT learns to use smaller memory more effectively than Transformer-XL. Additionally, the smaller memory size of RMT leads to reducing required GPU memory for running the model.

To force models to process longer recurrent dependencies the size of a segment is set to 50, so the number of recurrent steps increases. RMT with memory size 1 shows similar results to Transformer-XL with memory size 10. It is worth noting that Transformer-XL memory consists of hidden representations from all layers (in this case, it is $10 \times 16$ vectors) when RMT memory is only \texttt{memory\_size} vectors. Transformer-XL with memory size 50 and RMT with memory size 5 show similar perplexity values (see \cref{app:wt103}).

RMT could be combined with Tr-XL cache. In this case Tr-XL cache could be seen as short-term memory keeping the nearest context and RMT memory as long-term memory. Such combination leads to the best results on WikiText-103 improving over Tr-XL.

\begin{wraptable}{l}{0.5\linewidth}
\vspace{-0.2in}
\caption{\small \textbf{Language modeling on WikiText-103.} Average perplexity for the best performed variations of RMT models reported (see full results in \cref{app:wt103}). Underlined values show Tr-XL and RMT models with close results. RMT models with smaller memory sizes achieve similar scores to Tr-XL models with larger memory. Combination of cache with recurrent memory (Tr-XL + RMT) shows the best performance.}
\label{tab:wt103}
\begin{center}
\begin{small}
\begin{sc}
\fontsize{7}{8}
\selectfont 
\begin{tabular}{lccl}
\toprule
Model          & memory & segment len & $\text{ppl}_{\pm \text{std}}$ \\
\midrule
Tr-XL (paper)   &  150         &  150        & 24.0  \\
\midrule
Baseline        &  0           &  150          & 29.95 \tiny{$\pm$ 0.15} \\
MemTr           &  10          &  150         & 29.63 \tiny $\pm$ 0.06 \\
Tr-XL (ours)    &  150         &  150        & 24.12 \tiny $\pm$ 0.05 \\
\midrule
Tr-XL           &  25          &  150         & 25.57 \tiny $\pm$ 0.02 \\
Tr-XL           &  75          &  150         & \underline{24.68} \tiny $\pm$ 0.01 \\
RMT BPTT-3      &  10          &  150         & 25.04 \tiny $\pm$ 0.07 \\
RMT BPTT-2      &  25          &  150         & \underline{24.85} \tiny $\pm$ 0.31 \\
Tr-XL + RMT
&75+5 &  150        &  24.47 \tiny $\pm$ 0.05 \\
Tr-XL + RMT
&150+10&  150        & \textbf{23.99} \tiny $\pm$ 0.09 \\
\midrule
Baseline        &   0          &  50          & 39.05 \tiny $\pm$ 0.01 \\
Tr-XL           &   100         &  50         & \textbf{25.66} \tiny $\pm$ 0.01 \\
Tr-XL           &   50         &  50         & \underline{26.54} \tiny $\pm$ 0.01 \\
Tr-XL           &   25         &  50         & 27.57 \tiny $\pm$ 0.09\\
Tr-XL           &   10         &  50         & \dashuline{28.98} \tiny $\pm$ 0.11 \\
RMT BPTT-1      &   1          &  50          & \dashuline{28.71} \tiny $\pm$ 0.03 \\
RMT BPTT-3      &   10          &  50          & \underline{26.37} \tiny $\pm$ 0.01\\
\bottomrule
\end{tabular}
\end{sc}
\end{small}
\end{center}
\vskip -0.2in
\end{wraptable}

On enwik8 RMT models with memory size 5 and Transformer-XL with memory size 40 show similar results. Confirming that RMT learns to use smaller amounts of memory representation more effectively. All results for enwik8 dataset are shown in \cref{app:enwik8}. 

\begin{figure}[t]
\vspace{-0.1in}
\begin{center}
\begin{subfigure}[t]{0.49\linewidth}
    \centerline{\includegraphics[width=0.9\linewidth]{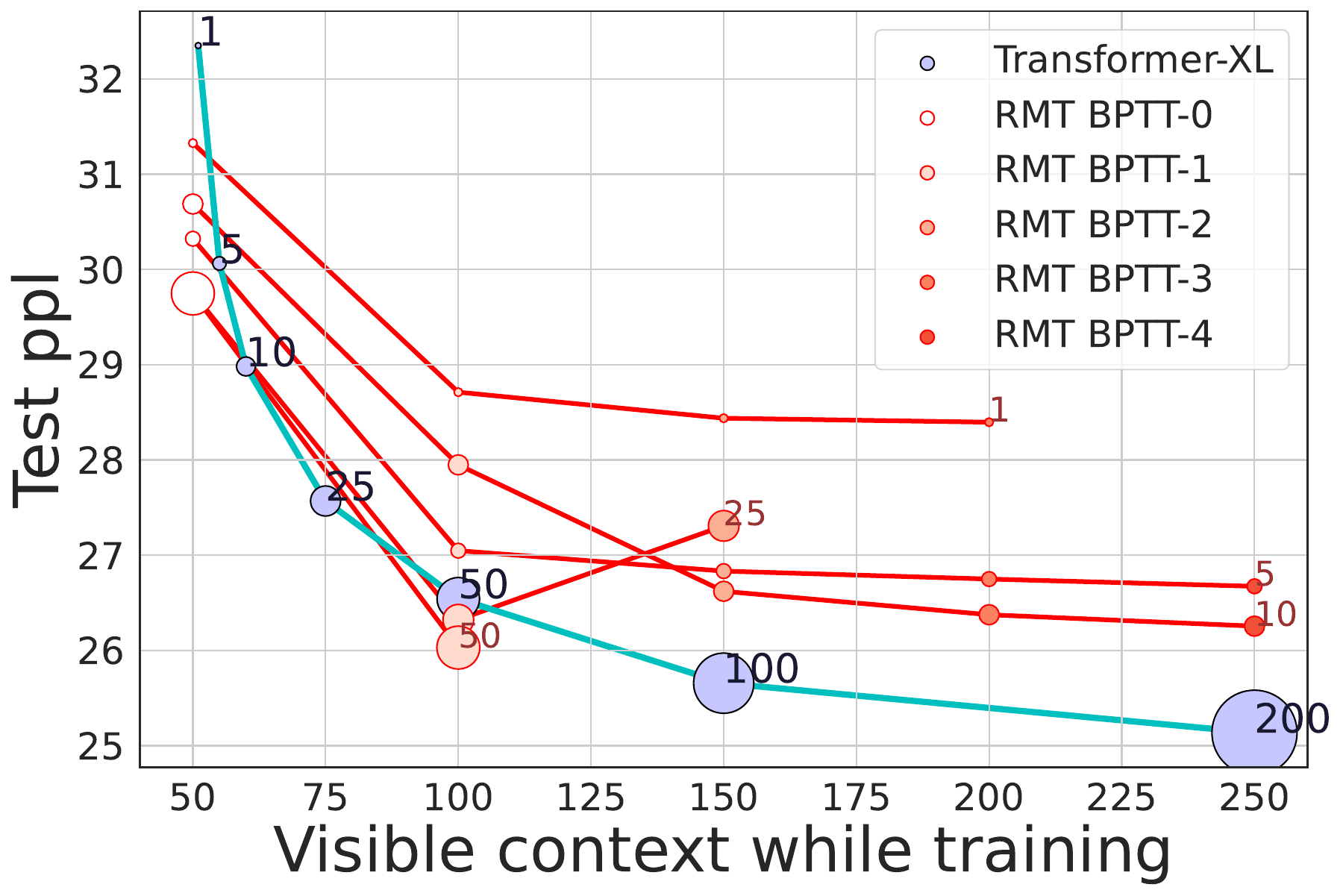}}
    \caption{}
    \label{fig:wt103_context}
\end{subfigure}\hfill
\begin{subfigure}[t]{0.49\linewidth}
    \centerline{\includegraphics[width=1\linewidth]{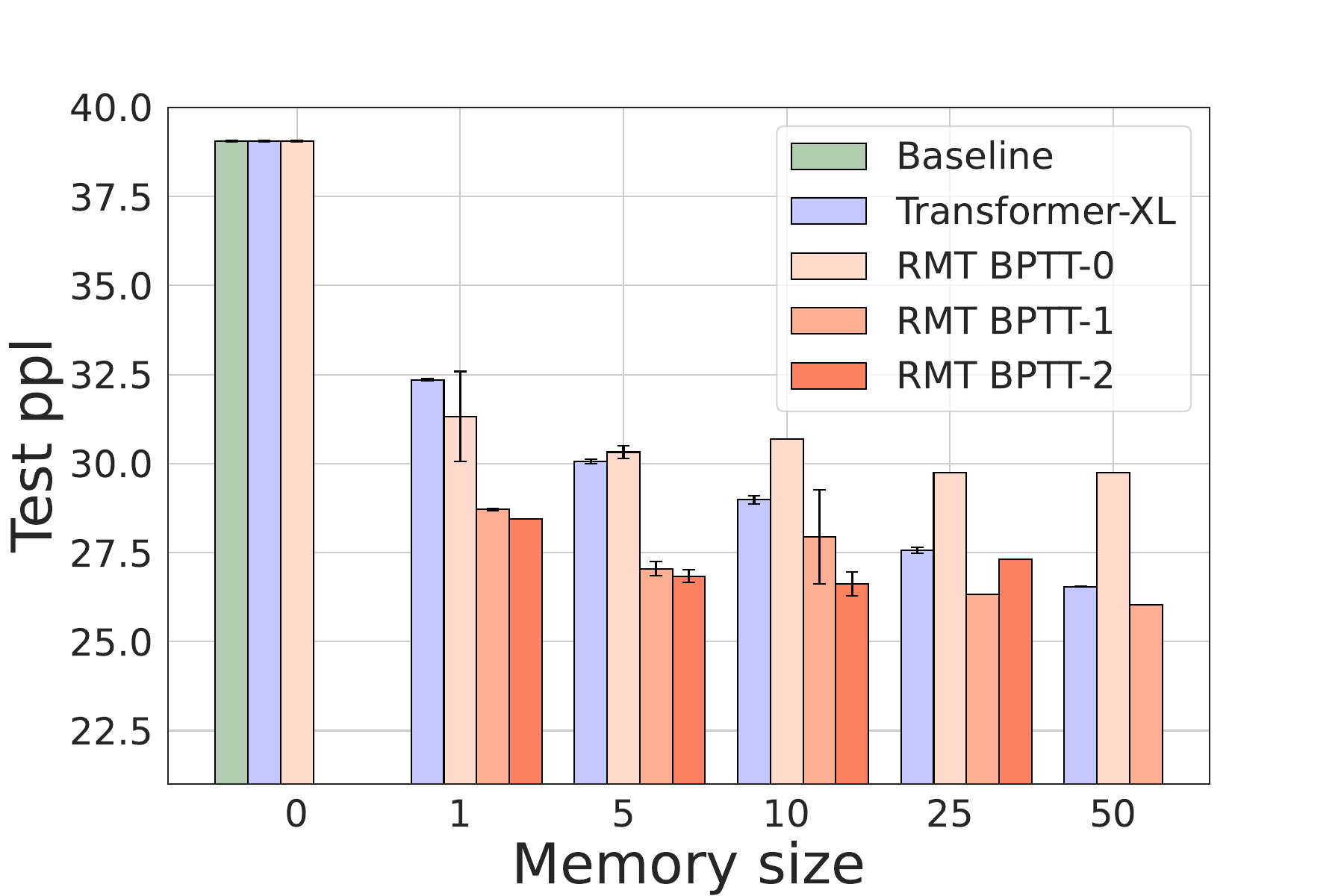}}
    \caption{}
    \label{fig:wt103_memory}
\end{subfigure}\hfill
\caption{\small \textbf{Deeper BPPT unrolling improves RMT scores on WikiText-103} (a) Visible context at training time can be increased by deeper BPTT unrolls for RMT or enlarging cache for Tr-XL. Larger visible context leads to lower perplexity for both models (marker size corresponds to memory size). (b) Recurrence improves performance of RMT compared to Tr-XL for the same memory sizes.}
\label{fig:rmt_lm_analysis}
\end{center}
\vskip -0.3in
\end{figure}

Recurrent Memory Transformer learns to make predictions depending on \texttt{\#BPTT\_unrolls} over previous segments $+1$ current segment. Transformer-XL does not use BPTT and relies only on \texttt{memory\_size} cached states and current segment making in total: \texttt{memory\_size} $+$ \texttt{segment\_length} tokens. In \cref{fig:wt103_context}, we compare RMT and Tr-XL according to the described value of visible context at training time.

RMT with a single memory vector could be trained to achieve lower perplexity as Transformer-XL with memory size 10. This means that RMT can learn to compress information from the previous observations better. Another observation is that RMT with memory sizes 10 and 25 performs only a bit weaker compared to Transformer-XL even when Transformer-XL has access to more non-compressed states (50, 100, 200) from previous segments. In general, training RMT with unrolling gradients in earlier segments drastically improves  scores thus showing the importance of BPTT training but, we observe instabilities and out-of-memory issues during RMT training for a larger memory sizes with deeper BPTT unrolls. 

RMT wins a lot when only one memory token is added but then the effect  from increasing memory size from 5 to 50 fades (\cref{fig:wt103_memory}). Still, RMT with memory size 5 have performance on par with Transformer-XL with cache 50, confirming that RMT learns to store more compact representations.  The results suggest that there is some optimal memory size for RMT to solve the task, and further increase does not add much. 

Proposed recurrent memory mechanism affects only input and gradient flows of the augmented core model. This might be an important advantage because the memory can be added to already pretrained model. Evaluation results for four memory augmented language models fine tuned for long text classification are presented in the \cref{tab:hyp}. Incorporation of 10 memory tokens in the input sequence of 512 allows to encode longer stretches of a text up to 2000 tokens and significantly improve metrics for the majority of models. Moreover, a combination of recurrent memory with RoBERTa-base results in state of the art performance for the Hyperpartisan news classification task \citep{kiesel2019semeval}. Interestingly, many competing models have input size of 4096 that is at least twice longer compared to RMT extended counterparts but still lag behind.   

\begin{wraptable}{r}{0.7\linewidth}
\vspace{-0.2in}
\caption{\small \textbf{Hyperpartisan news detection.} Models starting with RMT are taken from HuggingFace Transformers and augmented with 10 memory tokens and recurrence before fine-tuning. Train/valid/test split as in \citep{beltagy2020longformer} and metric is F1.}
\label{tab:hyp}
\begin{center}
\begin{small}
\begin{sc}
\fontsize{7}{8}
\selectfont 
\begin{tabular}{lcccc}
\toprule
\multirow{2}{*}{Model [input size]} & \multicolumn{4}{c}{number of segments} \\
& 1 & 2 & 3 & 4 \\
\midrule
Big Bird [4096] \citep{zaheer2020big} &  92.20 & & & \\
Longformer [4096] \citep{beltagy2020longformer} &  94.80 & & & \\
Graph-roberta [512x100] \citep{xu2021contrastive} &  96.15 & & & \\
ERNIE-DOC-Large [640] \citep{ding-etal-2021-ernie-doc} &  \underline{96.60} & & & \\
ERNIE-Sparse [4096] \citep{liu2022ernie} &  92.81 & & & \\
\midrule[0.1pt]
RMT bert-base-case [512] &  91.60 & 94.12 & 93.06 & 94.34 \\
RMT roberta-base [512] &  94.87 & \textbf{97.20} & \textbf{96.72} & \underline{\textbf{98.11}} \\
RMT deberta-v3-base [512] &  94.17 & 96.78 & 94.80 & 94.80 \\
RMT t5-base [512] &  \textbf{94.99} & 95.32 & 96.12 & 97.20 \\
\bottomrule
\end{tabular}
\end{sc}
\end{small}
\end{center}
\vskip -0.2in
\end{wraptable}

\begin{figure*}[!htp]
\begin{center}
\centerline{\includegraphics[width=\linewidth]{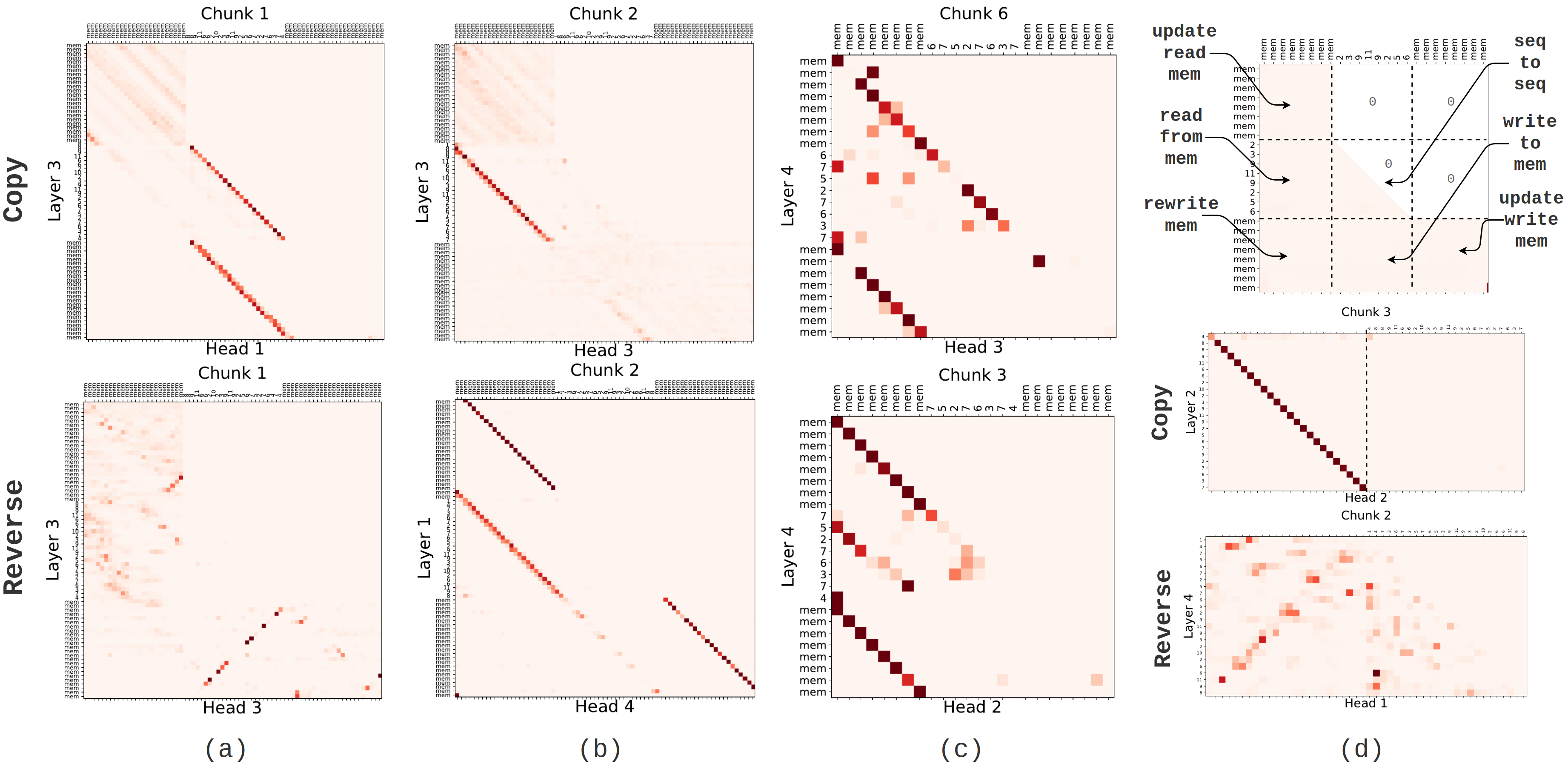}}
\caption{\small \textbf{Selected attention map patterns of memory models.} (color intensity corresponds to attention score) RMT with segment length=24, memory size=24 (a)  write to memory, (b) read from memory.  (c) RMT, segment length=8, memory size=8, rewrite from read memory to write memory. (d) Transformer-XL, segment length=24, memory size=24  read from the previous hidden states.}
\label{fig:mem_operations}
\end{center}
\vskip -0.35in
\end{figure*}

To get an understanding of memory operations, learned by RMT for algorithmic tasks we visualise attention maps for copy and reverse tasks (\cref{fig:mem_operations}). In each RMT attention map sequence tokens are preceded by read memory, located at the top left corner, and followed by write memory at the bottom right. Diagonal at the central part of the fig.\ref{fig:mem_operations}(a) (top)  shows classic attention of token sequence to itself, but the bottom diagonal represents the operation of writing of sequence tokens to memory in straight order. When completing reverse (fig.\ref{fig:mem_operations}(a) bottom)  the model learns to write the sequence to the memory in the reversed order, which is in line with common sense.

When it comes to reproducing the target sequence, the  model accesses memory (fig.\ref{fig:mem_operations}(b)) and writes to the output sequence. Another operation (fig.\ref{fig:mem_operations}(c)) is rewriting from read memory to write memory. It is commonly used by RMT in settings with larger number of segments to keep information about recent segments longer.

Transformer-XL mechanism of accessing memory (fig.\ref{fig:mem_operations}(d)) does not allow straightforward writing to memory without changing sequence token representations. Sequential reading from cache is represented by diagonals on Transformer-XL attention maps.
Using token representations as storage harms model performance in tasks with larger number of segments. 
For reverse task with 4 segments Transformer-XL with limited memory size 6 (\cref{app:mem_compression} \cref{fig:mem_compression}(b)) attempts to mix representations of tokens and read multiple symbols from one cached state in the next segments giving average accuracy of 0.8 on the target task. Despite having the same memory size, RMT manages to compress the whole segment in memory tokens (\cref{app:mem_compression} \cref{fig:mem_compression}(a)) and achieve mean accuracy 1.

Visualizations from \cref{fig:mem_operations} and \cref{app:mem_compression} \cref{fig:mem_compression} provide evidence to support our hypotheses that Tr-XL has to mix representations from previous and current segments in the same hidden states to pass information between segments. Also, visualizations show how memory tokens in RMT help mitigate such kind of mixing. RMT ability of sequence compression to memory is illustrated in \cref{app:alg_training_details} \cref{fig:limited_mem}. For copy with 6 segments RMT compresses and then reads the sequence of 12 tokens with just 6 memory tokens. For Transformer-XL decreasing memory size harms the accuracy score significantly with number of segments larger than 2.

\section{Conclusions}

In this paper we introduced Recurrent Memory Transformer a simple recurrent memory augmentation of Transformer model. RMT is implemented by extension of an input sequence with special global memory tokens and segment-level recurrence. Importantly, our method allows to learn more compact sequence representations and improve existing pretrained models without extensive additional compute, thus making practical machine learning applications more energy efficient and environmentally friendly.

In our experiments we compared RMT with Transformer baseline and Transformer-XL which is a well-known modification of Transformer for long sequences. RMT almost perfectly solves Copy, Reverse as well as quadratic equations tasks for sequences consisting of multiple segments outperforming Transformer-XL. It also demonstrates quality for associative retrieval task on par with Transformer-XL. As expected, baseline Transformer fails to solve these tasks for multi-segment settings.

RMT trained as a language model performs significantly ahead of Transformer baseline and shows quality metrics similar to Transformer-XL but for up to 10 times smaller memory size. Experimental results demonstrate that for fixed memory size backpropagating gradients for more segments improves performance of RMT. Proposed approach to memory augmentation is quite universal and might be easily applied to any pretrained transformer based model as demonstrated by achievement of state of the art results for long text classification task by fine tuning a combination of RoBERTa and RMT.

Analysis of attention maps suggests that better RMT performance can be related to more effective storage of input representations in dedicated memory tokens compared to mixing representations storage in Transformer-XL. RMT could be combined with Transformer-XL cache and improve the performance of both models.

Overall, results of the study show that dedicated memory storage and recurrence provided by Recurrent Memory Transformer make it a promising architecture for applications that require learning of long-term dependencies and general purpose in-memory processing, such as algorithmic tasks and reasoning. Furthermore, we believe that RMT could open the way for adding memory and recurrence to other models in the Transformer family.

\begin{ack}
This work was supported by a grant for research centers in the field of artificial intelligence, provided by the Analytical Center for the Government of the Russian Federation in accordance with the subsidy agreement (agreement identifier 000000D730321P5Q0002) and the agreement with the Moscow Institute of Physics and Technology dated November 1, 2021 No. 70-2021-00138.
\end{ack}


{
\small
\bibliographystyle{plainnat}
\bibliography{neurips_2022}
}

\newpage
\section*{Checklist}

\begin{enumerate}

\item For all authors...
\begin{enumerate}
  \item Do the main claims made in the abstract and introduction accurately reflect the paper's contributions and scope?
    \answerYes{}
  \item Did you describe the limitations of your work?
    \answerYes{We mention training instabilities and GPU RAM issues in~\cref{results}.}
  \item Did you discuss any potential negative societal impacts of your work?
    \answerNo{The proposed model and method do not have any specific impacts. All general negative societal impacts applicable to the field could be potentially relative.}
  \item Have you read the ethics review guidelines and ensured that your paper conforms to them?
    \answerYes{}
\end{enumerate}

\item If you are including theoretical results...
\begin{enumerate}
  \item Did you state the full set of assumptions of all theoretical results?
    \answerNA{}
        \item Did you include complete proofs of all theoretical results?
    \answerNA{}
\end{enumerate}

\item If you ran experiments...
\begin{enumerate}
  \item Did you include the code, data, and instructions needed to reproduce the main experimental results (either in the supplemental material or as a URL)?
    \answerYes{We include code, training scripts, and raw experimental data in the supplementary material. The supplemental materials would be published on github with the final version of the paper. Instructions for language modeling data\&experiments are taken from Tr-XL repo.}
  \item Did you specify all the training details (e.g., data splits, hyperparameters, how they were chosen)?
    \answerYes{See~\cref{sec:experiments},~\cref{app:training_details}, and provided supplementary material.}
        \item Did you report error bars (e.g., with respect to the random seed after running experiments multiple times)?
    \answerYes{All the key experiments results are reported with std. Furthermore, we provide raw experimental data in the supplementary materials.}
        \item Did you include the total amount of compute and the type of resources used (e.g., type of GPUs, internal cluster, or cloud provider)?
    \answerYes{We used different GPUs depending on the task: 1080Ti, V100, A100. We provide this information in~\cref{app:training_details} for each task.}
\end{enumerate}

\item If you are using existing assets (e.g., code, data, models) or curating/releasing new assets...
\begin{enumerate}
  \item If your work uses existing assets, did you cite the creators?
    \answerYes{We refer to the original Tr-XL code and Tr-XL paper. We use it for establishing baselines and setting our methods. See~\cref{sec:experiments}}
  \item Did you mention the license of the assets?
    \answerNo{Tr-XL license is Apache 2.0 and available at its github repo.}
  \item Did you include any new assets either in the supplemental material or as a URL?
    \answerYes{Our code is in the supplemental material and on GitHub: \url{https://github.com/booydar/LM-RMT}}
  \item Did you discuss whether and how consent was obtained from people whose data you're using/curating?
    \answerNo{We used publicly available Tr-XL code (Apache 2.0) and datasets.}
  \item Did you discuss whether the data you are using/curating contains personally identifiable information or offensive content?
    \answerNo{We use either synthetic data or datasets collected from the Wikipedia (Wikitext-103, enwik8).}
\end{enumerate}

\item If you used crowdsourcing or conducted research with human subjects...
\begin{enumerate}
  \item Did you include the full text of instructions given to participants and screenshots, if applicable?
    \answerNA{}{}
  \item Did you describe any potential participant risks, with links to Institutional Review Board (IRB) approvals, if applicable?
    \answerNA{}
  \item Did you include the estimated hourly wage paid to participants and the total amount spent on participant compensation?
    \answerNA{}
\end{enumerate}

\end{enumerate}


\newpage

\appendix

\section{Training details and additional results}
\label{app:training_details}

\subsection{Algorithmic tasks}
\label{app:alg_training_details}

Datasets were randomly generated by uniformly sampling tokens from dictionary into task sequences and generating targets accordingly to the tasks. After generation, datasets are fixed for all experiments.

Copy and reverse use sequences of sizes 24, 40, 120, 240, and 360, making total copy/reverse input length 48/72, 80/120, 240/360, 480/720, 720/1080. The associative retrieval task consists of 4 key-value pairs and one randomly selected key; the answer consists of one value. Train, validation and test sizes of copy 24, reverse 24 and associative retrieval datasets are 100000, 5000 and 10000.

Transformer-XL had the same cache size on training and validation to match RMT.

For training all models on copy and reverse, we used constant learning rate 1e-4 with reduction on plateau with decay factor of 0.5. Copy and reverse were solved by models with 4 layers and 4 heads, associative retrieval models had 6 layers and 4 heads.  Models with the same context size and memory size were trained for the same number of steps and the same training parameters. 

Experiments with sequence length 24 were conducted on a single Nvidia GTX 1080 Ti GPU from 1 hour to 2-3 days. Copy and reverse on longer sequence lengths were done on more powerful Tesla V100 using 1-3 devices with training time varying from 1 hour to 3-4 days.

\begin{figure}[htp]
\begin{center}
\centerline{\includegraphics[width=0.7\linewidth]{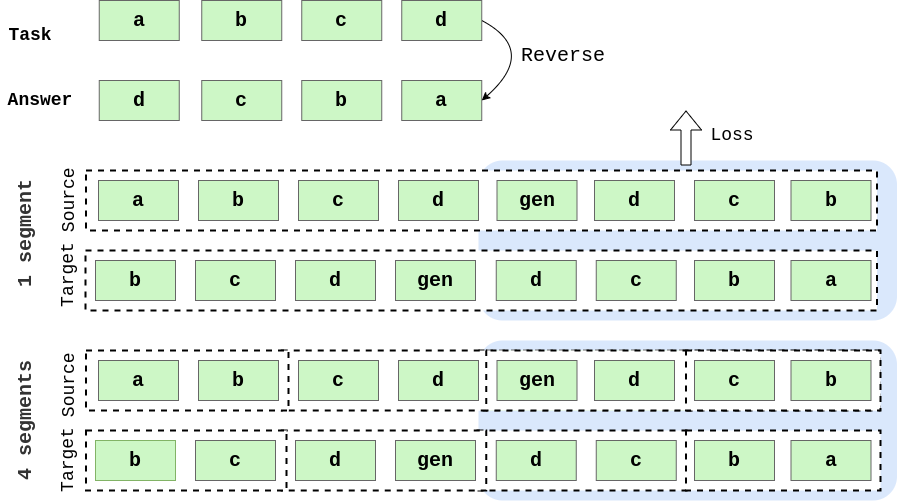}}
\caption{\small\textbf{Reverse task in one and four segments setting for decoder-only models.} Dotted lines show segment borders.
}
\label{fig:segments}
\end{center}
\end{figure}

\begin{figure}[h!]
\begin{center}
\centerline{\includegraphics[width=1\columnwidth]{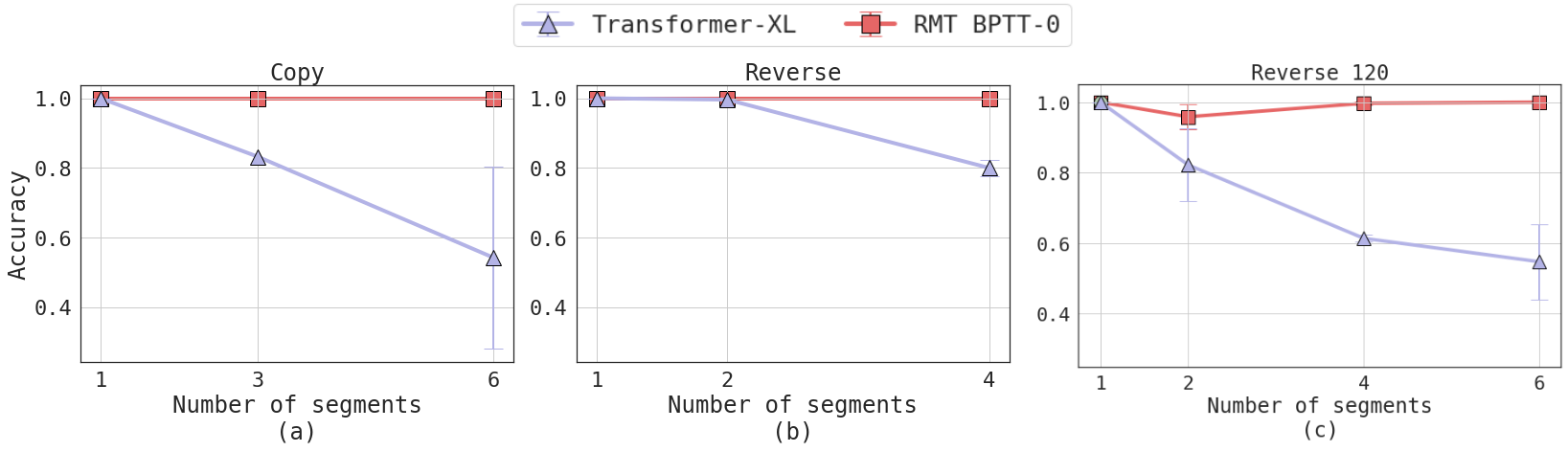}}
\caption{\small\textbf{RMT scales better with limited memory size.} Test set per-character accuracy on copy (a), reverse with a sequence length 24 (b) and 120 (c). Memory size is set to a half of the segment size. RMT solves the tasks almost perfectly with limited memory unlike Tr-XL.}
\label{fig:limited_mem}
\end{center}
\end{figure}

\subsection{Associative retrieval}
We used code for the task dataset generation from~\citep{NIPS2016_9f44e956}\footnote{\url{https://github.com/GokuMohandas/fast-weights/blob/539fb10e3c384d5f782af2560bf28631cd0eaa61/fw/data_utils.py}}.


\subsection{Quadratic equations}

This dataset consists of equations with integer coefficients and step-by-step solutions using the discriminant. Process of equation generation is started from uniformly sampling real roots $x_1, x_2$ from -100 to 100. The answer of an equation is represented as $x_1,x_2$. Next, we find the equation as multiplication of two parentheses $(x-x_1)(x-x_2)=0$, which is expanded to $x^2 - (x_1 + x_2)x + x_1 x_2 = 0$. Next, we multiply all coefficients by a random natural number $\alpha$ from 1 to 10. The final equation form is $\alpha x^2 - \alpha (x_1 + x_2)x + \alpha x_1 x_2 = 0$. 
A dataset sample is made of these stages in reversed order. We also provide a string with the discriminant calculation to help find the equation roots. 20 percent of equations in the dataset do not have real roots.

Example equation string:

\textit{-4*x\^{}2+392*x-2208=0}, 

solution string:

\textit{x\^{}2-98*x+552=0;D=98\^{}2-4*1*552=7396=86\^{}2;x=(98-86)/2=6;x=(98+86)/2=92} ,

and answer:

\textit{6,92}

Each solution step is tokenized on char level and padded to the length of 30 tokens. The total length of each training sample is 180, the dataset has 100000 training, 10000 validation and 20000 test samples.

For this task we used models with 6 layers, 6 heads and segment sizes 180 and 30. The training was performed with the same schedule as copy and reverse on a single GTX 1080 ti for 1-2 days. Memory size for RMT and Transformer-XL was chosen equal to the segment length.

\subsection{Enwik8}
\label{app:enwik8}

We verified our experimental setup by reproducing Transformer-XL results on enwik8 dataset (\cref{tab:enwik8}). We used 12-layer Baseline (Transformer), Transformer-XL, RMT in all enwik8 experiments. All results on enwik8 dataset are in \cref{tab:enwik8}. We used 2 NVIDIA A100 80Gb GPUs, training time varied from 10 to 30 hours depending on sequence length, memory size, and number of BPTT unrolls.

\begin{table}[h]
\caption{Test set bits-per-character on enwik8. Our experimental setup shows similar scores to the original paper \citep{dai2019transformerxl} with segment length 512.}
\label{tab:enwik8}
\vskip 0.0in
\begin{center}
\begin{small}
\begin{sc}
\begin{tabular}{lccl}
\toprule

Model          & memory & segment len  & $\text{bpc}_{\pm \text{std}}$ \\
\midrule
Tr-XL \citep{dai2019transformerxl}           &  512         &  512        & $1.06$ \\
Tr-XL (ours)           &  512         &  512        & $1.071$ \\
\midrule
Tr-XL         &  200         &  128        & 1.140 \\
Tr-XL         &  100         &  128        & 1.178 \\
Tr-XL         &  75         &  128         & 1.196 \\
Tr-XL           &  40         &  128       & 1.230 \tiny $\pm$ 0.001 \\
Tr-XL           &  20         &  128       & 1.261 \\
Tr-XL           &  10         &  128       & 1.283 \tiny $\pm$ 0.001 \\
RMT BPTT-1      &  5         &  128          & 1.241 \tiny $\pm$ 0.002 \\
RMT BPTT-2      &  5         &  128          & 1.231 \tiny $\pm$ 0.002 \\
RMT BPTT-1      &  10         &  128         & 1.240 \tiny $\pm$ 0.006\\
RMT BPTT-2      &  10         &  128         & 1.228 \tiny $\pm$ 0.003\\
RMT BPTT-0        &  20         &  128         &  1.301\\
RMT BPTT-1      &  20         &  128         & 1.229\\
RMT BPTT-2      &  20         &  128         & 1.222\\

\bottomrule
\end{tabular}
\end{sc}
\end{small}
\end{center}
\vskip -0.1in
\end{table}

\subsection{WikiText-103}
\label{app:wt103}

We used 16-layer models in all experiments on WikiText-103 dataset. Training hyperparameters were used from ~\citep{dai2019transformerxl} and authors PyTorch scripts\footnote{\url{https://github.com/kimiyoung/transformer-xl}}. All results on WikiText-103 dataset are in \cref{tab:wt103_full}. In most of the WikiText-103 experiments, we used 2 NVIDIA A100 80Gb GPUs, training time varied from 10 to 30 hours depending on sequence length, memory size, and number of BPTT unrolls.
All models except the ones noted with 2x steps were trained for 200k batches. Transformer-XL did not benefit from longer training unlike the Tr-XL + RMT model. For training the combined model we used an auxiliary loss for memory tokens, it was added to the main loss with a multiplier of $0.01$. We set a new fixed special token to be predicted from memory as target in the auxiliary loss.

\begin{table}[]
\caption{Test set perplexity on WikiText-103. All experiments with RMT and Tr-XL models.}
\label{tab:wt103_full}
\vskip 0.0in
\begin{center}
\begin{small}
\begin{sc}
\begin{tabular}{lccl}
\toprule
Model          & memory & segment len & $\text{ppl}_{\pm \text{std}}$ \\
\midrule
Baseline        &  0           &  150          & 29.95 \tiny{$\pm$ 0.15} \\
MT           &  10          &  150         & 29.63 \tiny $\pm$ 0.06 \\
MT           &  25          &  150         & 29.67 \tiny $\pm$ 0.03 \\
MT           &  75          &  150         & 29.69 \tiny $\pm$ 0.02 \\
MT           &  150          &  150         & 29.82 \tiny $\pm$ 0.35 \\
Tr-XL (paper)   &  150         &  150        & 24.0  \\
Tr-XL (ours)    &  150         &  150        & 24.12 \tiny $\pm$ 0.05 \\
Tr-XL (ours) 2x steps   &  150 &  150        & 24.67  \\
Tr-XL           &  75          &  150         & 24.68 \tiny $\pm$ 0.01 \\
Tr-XL 2x steps  &  75          &  150         & 24.49  \\
Tr-XL           &  25          &  150         & 25.57 \tiny $\pm$ 0.02 \\
RMT BPTT-0      &  10          &  150         & 26.85 \tiny $\pm$ 0.02 \\
RMT BPTT-1      &  10          &  150         & 25.92 \tiny $\pm$ 1.07 \\
RMT BPTT-2      &  10          &  150         & 25.32 \tiny $\pm$ 0.61 \\
RMT BPTT-3      &  10          &  150         & 25.04 \tiny $\pm$ 0.07 \\
RMT BPTT-0      &  25          &  150         & 29.73 \\
RMT BPTT-1      &  25          &  150         & 24.91 \\
RMT BPTT-2      &  25          &  150         & 24.85 \tiny $\pm$ 0.31 \\
Tr-XL + RMT BPTT-3 & 70 + 5    &  150         & 24,53 \\ 
Tr-XL + RMT BPTT-3 & 75 + 5    &  150         & 24,47 \tiny $\pm$ 0.05 \\ 
Tr-XL + RMT BPTT-0 & 140 + 10  &  150         & 24,25 \\ 
Tr-XL + RMT BPTT-1 & 150 + 10  &  150         & 24,30 \tiny $\pm$ 0.09 \\ 
Tr-XL + RMT BPTT-3 2x steps & 150 + 10  &  150         & 23,99 \tiny $\pm$ 0.09 \\ 
\midrule
Baseline        &   0          &  50          & 39.05 \tiny $\pm$ 0.01 \\
Tr-XL           &   200         &  50         & 25.14 \\
Tr-XL           &   100         &  50         & 25.66 \tiny $\pm$ 0.01 \\
Tr-XL           &   50         &  50         & 26.54 \tiny $\pm$ 0.01 \\
Tr-XL           &   25         &  50         & 27.57 \tiny $\pm$ 0.09\\
Tr-XL           &   10         &  50         & 28.98 \tiny $\pm$ 0.11 \\
Tr-XL           &   5         &  50         & 30.06 \tiny $\pm$ 0.07 \\
Tr-XL           &   1          &  50         & 32.35 \tiny $\pm$ 0.03 \\
RMT BPTT-0      &   1          &  50          & 31.33 \tiny $\pm$ 1.26 \\
RMT BPTT-1      &   1          &  50          & 28.71 \tiny $\pm$ 0.03 \\
RMT BPTT-2      &   1          &  50          & 28.44 \\
RMT BPTT-3      &   1          &  50          & 28.40 \tiny $\pm$ 0.03 \\
RMT BPTT-0      &   5          &  50          & 30.32 \tiny $\pm$ 0.18\\
RMT BPTT-1      &   5          &  50          & 27.05 \tiny $\pm$ 0.20\\
RMT BPTT-2      &   5          &  50          & 26.83 \tiny $\pm$ 0.18\\
RMT BPTT-3      &   5          &  50          & 26.75 \tiny $\pm$ 0.26\\
RMT BPTT-4      &   5          &  50          & 26.67 \tiny $\pm$ 0.03\\
RMT BPTT-0      &   10          &  50          & 30.69 \tiny $\pm$ 0.01\\
RMT BPTT-1      &   10          &  50          & 27.95 \tiny $\pm$ 1.32\\
RMT BPTT-2      &   10          &  50          & 26.62 \tiny $\pm$ 0.34\\
RMT BPTT-3      &   10          &  50          & 26.37 \tiny $\pm$ 0.01\\
RMT BPTT-4      &   10          &  50          & 26.25 \tiny $\pm$ 0.19\\
RMT BPTT-0      &   25          &  50          & 29.75 \\
RMT BPTT-1      &   25          &  50          & 26.32 \\
RMT BPTT-2      &   25          &  50          & 27.31 \\
RMT BPTT-0      &   50          &  50          & 29.75 \\
RMT BPTT-1      &   50          &  50          & 26.03 \\

\bottomrule
\end{tabular}
\end{sc}
\end{small}
\end{center}
\vskip -0.1in
\end{table}

\clearpage
\section{Operations with Memory}
\label{app:mem_compression}

\begin{figure}[ht]
\vskip 0.2in
\begin{center}
\centerline{\includegraphics[width=0.7\columnwidth]{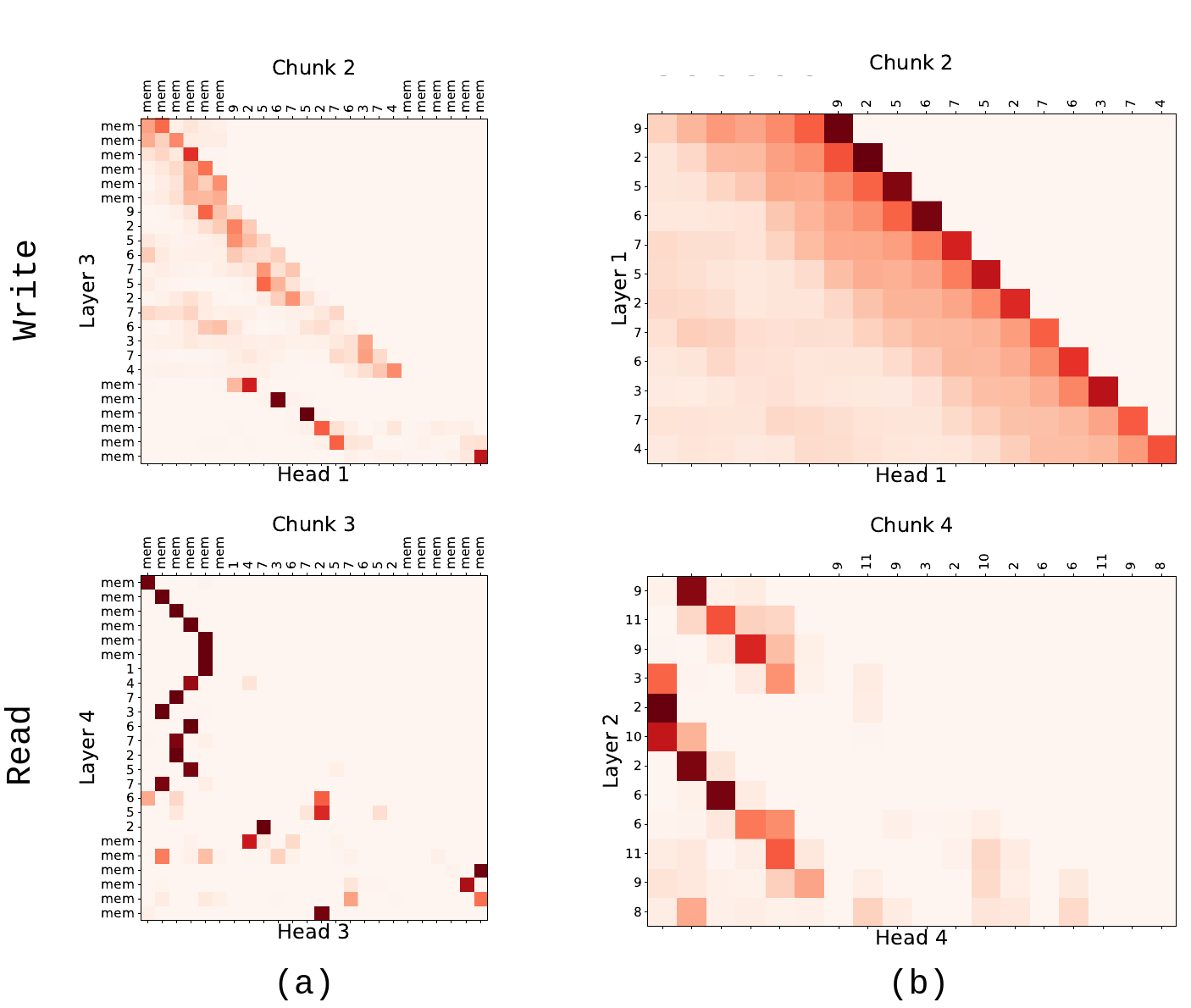}}
\caption{Approaches to compression and decompression of sequence with length 12 and memory with size 6. (a) - RMT, (b) - Transformer-XL. Tr-XL mixes representations of tokens and reads multiple symbols from one cached state. RMT manages to compress the whole segment into memory tokens.}
\label{fig:mem_compression}
\end{center}
\vskip -0.2in
\end{figure}

\end{document}